\documentclass{sig-alternate-ipsn13}

\usepackage{graphicx}
\usepackage{subfloat}
\usepackage{subfigure}
\usepackage{amsmath}
\usepackage{caption}
\usepackage{textpos}
\usepackage{mathtools}
\usepackage{tabularx}
\usepackage{multicol}
\usepackage{url}
\usepackage{breakurl}

\begin{document}

\title{An unsupervised spatiotemporal graphical modeling approach to anomaly detection in distributed CPS\titlenote{This material is based upon work supported by the National Science Foundation under Grant No. CNS-1464279.}}
\numberofauthors{1} 
%
\author{
\alignauthor
Chao Liu, Sambuddha Ghosal, Zhanhong Jiang, and Soumik Sarkar\titlenote{Corresponding author.}\\
       \affaddr{Department of Mechanical Engineering, Iowa State University, Ames, IA 50011, USA}\\
       \email{\{cliu5, sghosal, zhjiang, soumiks\}@iastate.edu}
}
\maketitle
\begin{abstract}
  Modern distributed cyber-physical systems (CPSs) encounter a large variety of physical faults and cyber anomalies and in many cases, they are vulnerable to catastrophic fault propagation scenarios due to strong connectivity among the sub-systems. This paper presents a new data-driven framework for system-wide anomaly detection for addressing such issues. The framework is based on a spatiotemporal feature extraction scheme built on the concept of symbolic dynamics for discovering and representing causal interactions among the subsystems of a CPS. The extracted spatiotemporal features are then used to learn system-wide patterns via a Restricted Boltzmann Machine (RBM). The results show that: (1) the RBM free energy in the off-nominal conditions is different from that in the nominal conditions and can be used for anomaly detection; (2) the framework can capture multiple nominal modes with one graphical model; (3) the case studies with simulated data and an integrated building system validate the proposed approach.
\end{abstract}

\section{Introduction}
With the advent ubiquitous sensing, advanced computation and strong connectivity, modern distributed cyber-physical systems (CPSs) such as power plants~\cite{kesler2011vulnerability}, integrated buildings~\cite{kleissl2010cyber}, transportation networks~\cite{work2008impacts} and power-grids~\cite{sridhar2012cyber} have shown tremendous potential of increased efficiency, robustness and resilience - a fact that is reflected in  great interest in CPS research shown by the U.S. government~\cite{kramer2014white}, the EU~\cite{fitzgerald2013model}, other countries around the world~\cite{sztipanovits2012strategic} and industry in general~\cite{bradley2013embracing}. However, to realize such potential, effective modeling and analysis tools have to be developed for CPSs that are scalable, robust, flexible and adaptive. Most of the current approaches lack in these qualities as they heavily depend on domain knowledge based rules and first-principle based models that need meticulous calibration and validation.

From the perspective of performance monitoring and diagnostics of distributed CPSs, the technical challenges arise from a large number of subsystems that are highly interactive~\cite{sanislav2012cyber} and operate in diverse modes. Although physics-based modeling of individual small sub-systems, developing models that capture all possible complex interactions often become intractable. Data-driven modeling can potentially alleviate such issues~\cite{CZD11}. However, most of such methods need examples of nominal and all possible anomalous conditions (e.g., cyber attacks and physical faults) which is intractable for real life systems. Therefore, anomaly detection approaches should have (i) the potential to recognize most of the operating modes without anomaly as nominal, and (ii) an unsupervised learning ability to distinguish the (possibly unforeseen) anomalies from the nominal modes. Moreover, physical space generates mostly continuous temporal information from sensors and actuators and the cyber space generates mostly discrete event-driven data while processing physical information. Such disparity in fundamental properties and nature of information space drives most of the current approaches to treat cyber and physical spaces separately for modeling and analysis (a detail survey can be found in~\cite{KST14}).

In this context, this work presents a data driven framework for system-wide anomaly detection in distributed CPSs, where a spatiotemporal feature extraction scheme is built on the concept of symbolic dynamics~\cite{RRSY09} for discovering and representing causal interactions between the subsystems. Symbolic Dynamic Filtering (SDF), as a data driven modeling of complex systems, has advantages in describing different types of data with a uniform representation, named as data abstraction. Abstraction involves pre-processing and data space partitioning for relevant variables (e.g., sensor time-series, event logs) of the system~\cite{sarkar2009generalization}, where such a representation helps modeling cyber and physical sub-systems together. Features captured by SDF are used in formation of spatiotemporal pattern network (STPN) \cite{jiang2015under}, a recently proposed causal graphical modeling concept. The causal modeling is followed by unsupervised learning of various system-level nominal patterns~\cite{fiore2013network}. Upon learning such models, the paper develops inference schemes for detection of low probability events or anomalies.

\section{Background and preliminaries}

\subsection{Spatiotemporal pattern network (STPN)}
\label{secSDF}
Symbolic dynamic filtering (SDF) has been recently shown to be extremely effective for extracting key textures from time-series data for anomaly detection and pattern classification~\cite{RRSY09}. The core idea is that a symbol sequence (i.e., discretized time-series) emanated from a process can be approximated as a Markov chain of order $D$ (also called depth), named as $D$-Markov machine \cite{sarkar2014sensor} that captures key behavior of the underlying process. The discretization or symbolization process is called partitioning~\cite{RRSY09}. There are various approaches proposed in the literature, depending on different objective functions \cite{Soumalya2015phd}, such as uniform partitioning (UP), maximum entropy partitioning (MEP), maximally bijective discretization (MBD) \cite{sarkar2013maximally}. This study uses simple uniform partitioning. The $D$-Markov machine is essentially a probabilistic finite state automaton (PFSA) that can be described by states (representing various parts of the data space) and probabilistic transitions among them that can be learnt from data. Related definitions of deterministic finite state automaton (DFSA), PFSA, $D$-Markov machine, $xD$-Markov machine and the learning schemes can be found in~\cite{sarkar2014sensor}.

With this setup, a spatiotemporal pattern network (STPN) is defined below.

\begin{figure}[h]
  \centering
  \includegraphics[scale=0.39]{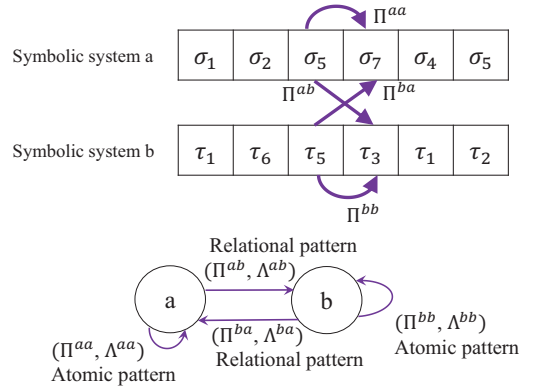}
  \caption{\emph{Extraction of atomic and relational patterns (using $D$-Markov and $xD$-Markov machines respectively and $D = 1$, i.e., states and symbols are equivalent) to characterize individual sub-system behavior and interaction behavior among different sub-systems.
  }}
  \label{figDemoStpn}\vspace{-10pt}
\end{figure}

\textbf{Definition}. A PFSA based STPN is a 4-tuple $W_{D} \equiv ( Q^{a}, \Sigma^{b}, \Pi^{ab}, \Lambda^{ab})$: (a, b denote nodes of the STPN)
\begin{enumerate}
  \item $Q^{a}=\{q_{1}, q_{2}, \cdots, q_{|Q^{a}|}\}$ is the state set corresponding to symbol sequences ${S^{a}}$;
  \item $\Sigma^{b}=\{\sigma_{0}, \cdots, \sigma_{|\Sigma^{b}|-1}\}$ is the alphabet set of symbol sequence ${S^{b}}$;
  \item $\Pi^{ab}$ is the symbol generation matrix of size $|Q^{a}| \times |\Sigma^{b}|$, the $ij^{th}$ element of $\Pi^{ab}$ denotes the probability of finding the symbol $\sigma_{j}$ in the symbol string  ${s^{b}}$ while making a transition from the state $q_{i}$ in the symbol sequence ${S^{a}}$; while self-symbol generation matrices are called atomic patterns (APs) i.e., when $a=b$, cross-symbol generation matrices are called relational patterns (RPs) i.e., when $a \neq b$.
  \item $\Lambda^{ab}$ denotes a metric that can represent the importance of the learnt pattern (or degree of causality) for $a \rightarrow b$ which is a function of $\Pi^{ab}$.
\end{enumerate}
An illustration of STPN is shown in Fig. \ref{figDemoStpn}.

Information based criteria are often used to compute a metric such as $\Lambda^{ij}$, e.g., transfer entropy \cite{wibral2011transfer} and mutual information \cite{solo2008causality}.
Detailed description of mutual information based causality metric in the context of APs and RPs can be found in~\cite{jiang2015under,sarkar2014sensor}.

\begin{figure*}[t]
  \centering
  \includegraphics[scale=0.48]{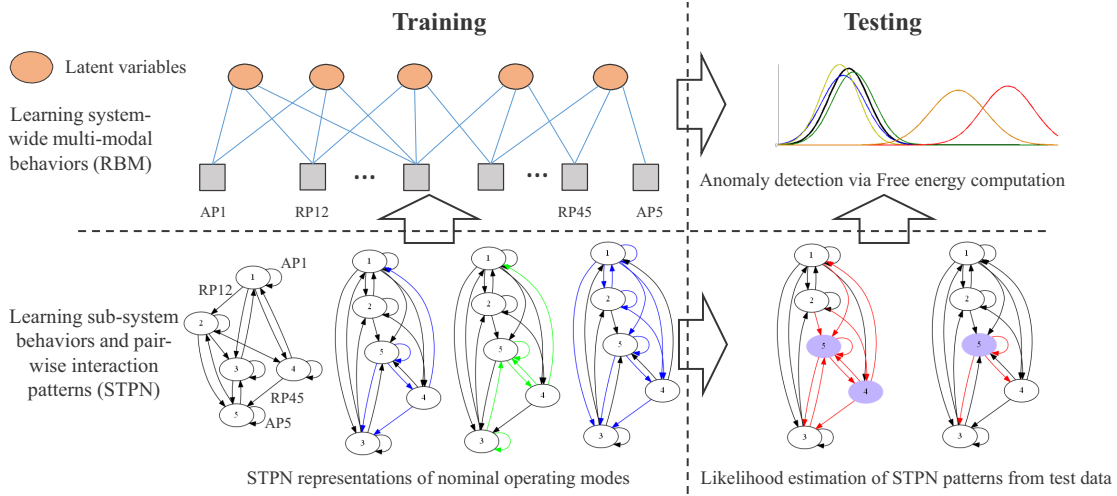}
  \caption{\emph{A data driven framework for anomaly detection in distributed CPS (i) Spatiotemporal graphical modeling learn nominal behavior of subsystems and their interactions in various operation modes, and system-wide patterns are learnt by an RBM, (ii) distribution of free energy is used to detect low probability events or anomalies}}
  \label{figFramework}\vspace{-10pt}
\end{figure*}

\subsection{Restricted Boltzmann Machine (RBM)}\label{sec:rbm}
RBM has grabbed a lot of recent attention in the Deep Learning community \cite{HS06,RB08}
for unsupervised feature extraction. The basic structure of RBM is shown in unsupervised learning layer in Fig.~\ref{figFramework} (top-left corner). As an energy based model \cite{HS06}, weights and biases are learnt so that the feature configurations observed during nominal operation of the system gets low energy (or high probability). Consider a system state that is described by a set of visible variables $\textbf{v} = (v_1, v_2, \cdots, v_D)$ and a set of hidden (latent) variables $\textbf{h} = (h_1, h_2, \cdots, h_F)$. The variables can be binary or real-valued depending on the need. Now, each joint configuration of these variables determines a particular state of the system and an energy value $E(\textbf{v}, \textbf{h})$ is associated with it. The energy values are functions of the weights of the links between the variables (for RBM, internal links within the visible variables and the hidden variables are not considered) and bias terms related to the variables.

With this setup, the probability of a state $P(\textbf{v}, \textbf{h})$ depends only on the energy of the configuration (\textbf{v}, \textbf{h}) and follows the Boltzmann distribution
\begin{equation}
  P(\textbf{v}, \textbf{h}) = \frac{e^{-E(\textbf{v}, \textbf{h})}}{\sum_{\textbf{v}, \textbf{h}}e^{-E(\textbf{v}, \textbf{h})}}
\end{equation}
Typically, during training, weights and biases are obtained via maximizing likelihood of the training data.

\section{Proposed methodology}\label{graphUlearn}
The proposed data-driven framework for system-wide anomaly detection is shown in Fig.~\ref{figFramework}. During training, the steps of learning the STPN+RBM model are:
\begin{enumerate}
\item Learn APs and RPs (individual node behaviors and pair-wise interaction behaviors) from the multivariate training symbol sequences \vspace{-4pt}
\item Consider short symbol subsequences from the training sequences and evaluate $\Lambda^{ij}\ \forall i,j$ for each short subsequence \vspace{-4pt}
\item For one subsequence, based on a user-defined threshold on $\Lambda^{ij}$, assign state $0$ or $1$ for each AP and RP; thus every subsequence leads to a binary vector of length $L$, where $L = \#AP + \#RP$ \vspace{-4pt}
\item An RBM is used for modeling system-wide behavior with nodes in the visible layer corresponding to APs and RPs \vspace{-4pt}
\item The RBM is trained using binary vectors generated from training subsequences
\end{enumerate}

\subsection{Training STPN+RBM model of a CPS}
Consider a multivariate training time series (nominal operation data in the present context), $X=\{X^{\mathbb{A}}(t)$, $t \in \mathbb{N}$, $\mathbb{A}=1,2,\cdots, f \}$, where $f$ is the number of variables or dimension of the time series. At first, symbolization and PFSA learning are performed to extract atomic and relational patterns for the corresponding STPN with $f$ vertices and $f^2$ edges. In this case, let the set of symbol sequences be $S=\{S^{\mathbb{A}}\}$. Then, we define a short subsequence, $\tilde{X}=\{\tilde{X}^{\mathbb{A}}(t)$, $t \in \mathbb{N}^{*}$, $\mathbb{A}=1,2,\cdots, f \}$, where $\mathbb{N}^{*}$ is a subset of $\mathbb{N}$. Essentially, different (possibly overlapping) time windows (represented by $t \in \mathbb{N}^{*}$) extracted from the overall training data set can be considered as the short subsequences. Similar to the previous notation, a set of symbolic subsequences for one time window is denoted as $\tilde{S}=\{\tilde{S}^{\mathbb{A}}\}$.

The next step is to compute $\Lambda^{ij}\ \forall i,j$ for every short subsequence extracted from the entire time series. Although information theoretic measures (as mentioned above) can be good candidates, estimation of such metrics need a large number of data points and hence, can be prohibitive at the anomaly detection phase. In this paper, we propose a new measure of $\Lambda^{ij}\ \forall i,j$ based on the statistical inference strategy developed in~\cite{sarkar2013symbolic,akintayo2015symbolic}.

Computation of this metric involves two phases, namely (i) modeling phase and (ii) inference phase.

\vspace{2pt}
\textbf{Modeling Phase:} The entire training data is considered in the context of the STPN, $W_{D} \equiv ( Q^{a}, \Sigma^{b}, \Pi^{ab}, \Lambda^{ab})$. As the entire set of symbol sequences is denoted by $S$, let the set of state sequences (obtained by applying the depth $D$ on the symbol sequence) be denoted as $\mathcal{Q}=\{\mathcal{Q}^a,\ a=1,2,\cdots, f \}$. Recall, a pattern $\Pi^{ab}$ depends on the state sequence $\mathcal{Q}^a$ and the symbol sequence $S^b$.
%
%
With this setup, each row of $\Pi^{ab}$ is treated as a random vector. For the $m^{th}$ row, a prior probability density function $f_{\Pi_{m}^{ab}|\{\mathcal{Q}^{a}, S^b\}}$ for the random vector $\Pi_{m}^{ab}$, conditioned on the joint state-symbol sequence $\{\mathcal{Q}^{a},S^b\}$ following the Dirchlet distribution is described below
\begin{equation}
f_{\Pi_{m}^{ab}|\{\mathcal{Q}^{a},S^b\}} = \frac{1}{B(\alpha_{m}^{ab})}\prod_{n=1}^{|\Sigma^{b}|}{(\theta_{mn}^{ab})^{\alpha_{mn}^{ab}-1}}
\label{dirchlet}
\end{equation}
where $\theta_{m}^{ab}$ is a realization of the random vector $\Pi_{m}^{ab}$, namely
\begin{equation*}
\theta_{mn}^{ab}=
  \begin{bmatrix}
    \theta_{m1}^{ab} \quad \theta_{m2}^{ab} \quad \cdots \quad \theta_{m|\Sigma^{b}|}^{ab}
  \end{bmatrix}
\end{equation*}
and the normalizing constant is
\begin{equation}
B(\alpha_{m}^{ab}) \triangleq \frac{\prod\limits_{n=1}^{|\Sigma^{b}|}{\Gamma(\alpha_{mn}^{ab})}} {\Gamma(\sum\limits_{n=1}^{|\Sigma^{b}|}{\alpha_{mn}^{ab}})}
\label{alpha1}
\end{equation}
where $\alpha_{mn}^{ab} \triangleq [\alpha_{m1}^{ab} \ \alpha_{m2}^{ab} \ \cdots \ \alpha_{m|\Sigma^{b}|}^{ab}]$ with $\alpha_{mn}^{ab}=N_{mn}^{ab}+1$ and $N_{mn}^{ab}$ is the number of times the symbol $\sigma^{b}_n \in \Sigma^{b}$ is emanated after the state $q^{a}_m \in Q^{a}$, i.e.,
\begin{equation}
N_{mn}^{ab} \triangleq |\{(\mathcal{Q}^{a}(k),S^{b}(k+1)): S^{b}(k+1) = \sigma^{b}_n\ | \ \mathcal{Q}^{a}(k) = q^a_m \}| 
\label{eqNmn1}
\end{equation}
where $\mathcal{Q}^{a}(k)$ is the $k^{th}$ state in the state sequence $\mathcal{Q}^{a}$ and $S^{b}(k+1)$ is the $(k+1)^{th}$ symbol in the symbol sequence $S^{b}$.

It follows from Eq. \ref{alpha1} that
\begin{align}
\begin{split}
B(\alpha_{m}^{ab}) = \frac{\prod\limits_{n=1}^{|\Sigma^{b}|}{\Gamma(N_{mn}^{ab}+1)}}{\Gamma(\sum_{n=1}^{|\Sigma^{b}|}{N_{mn}^{ab}+|\Sigma^{b}|})}
                   = \frac{\prod\limits_{n=1}^{|\Sigma^{b}|}{(N_{mn}^{ab})!}}{(N_{m}^{ab}+|\Sigma^{b}|-1)!}
\label{alpha2}
\end{split}
\end{align}
by using the relation $\Gamma(n)=(n-1)!$. 

With the Markov property of the learned PFSA, the row-vectors, $\{ \Pi_{m}^{ab}, \ m=1,2,\cdots,|Q^{a}| \}$, are statistically independent of each other. Therefore, it follows Eqs. \ref{dirchlet} and \ref{alpha2} that the prior joint density $f_{\Pi_{m}^{ab}|\{\mathcal{Q}^{a},S^b\}}$ of the probability morph matrix $\Pi^{ab}$, conditioned on the joint state-symbol sequences $\{\mathcal{Q}^{a},S^b\}$, is given by
\begin{align}
\begin{split}
f_{\Pi^{ab}|\{\mathcal{Q}^{a},S^b\}}(\theta^{ab}|\{\mathcal{Q}^{a},S^b\})\\
             =\prod_{m=1}^{|Q^{a}|} f_{\Pi_m^{ab}|\{\mathcal{Q}^{a},S^b\}} (\theta_{m}^{ab}|\{\mathcal{Q}^{a},S^b\})\\
            =\prod_{m=1}^{|Q^{a}|}{(N_{m}^{ab}+|\Sigma^{b}|-1)!}\prod_{n=1}^{|\Sigma^{b}|}\frac{(\theta_{m}^{ab})^{N_{mn}^{ab}}}{(N_{mn}^{ab})!}
\label{prior}
\end{split}
\end{align}
where $\theta^{ab}=[(\theta_{1}^{ab})^{T} \ (\theta_{2}^{ab})^{T} \ \cdots \  (\theta_{|Q^{a}|}^{ab})^{T} \ ] \in [0,1]^{|Q^{a}|\times|\Sigma^{b}|}$.

\vspace{2pt}
\textbf{Inference Phase:} After modeling with the entire set of training sequences, the goal of the inference phase is to compute the metric $\Lambda^{ab}(\tilde{\mathcal{Q}})$ for a given short subsequence described by $\tilde{\mathcal{Q}}$ and $\tilde{S}$. The value of this metric suggests the importance of the pattern $\Pi^{ab}$ or the degree of causality in $a \rightarrow b$ as evidenced by the short subsequence. In this context, we consider
\begin{equation}
\Lambda^{ab}(\tilde{\mathcal{Q}}, \tilde{S}) \propto Pr(\{\tilde{\mathcal{Q}}^{a},\tilde{S}^b\}|\Pi^{ab})
\end{equation}
where the probability of the joint state-symbol subsequence is a product of independent multinominal distributions given that the exact morph matrix is known.

\begin{align}
\begin{split}
Pr(\{\tilde{\mathcal{Q}}^{a},\tilde{S}^b\}|\Pi^{ab}) = \prod_{m=1}^{|Q^{a}|}{(\tilde{N}_{m})!} \prod_{n=1}^{|\Sigma^{b}|}\frac{(\Pi_{mn}^{ab})^{\tilde{N}_{mn}}}{(N_{mn}^{ab})!}
\label{post}
\end{split}
\end{align}
where, the definition of $\tilde{N}_{mn}^{ab}$ is similar to $N_{mn}^{ab}$ in the context of the short subsequence.

With similar derivation in \cite{sarkar2013symbolic}, the metric $\Lambda^{ab}(\tilde{\mathcal{Q}}, \tilde{S})$ can be obtained as follows
\begin{align}
\begin{split}
\Lambda^{ab}(\tilde{\mathcal{Q}}, \tilde{S}) =
     K \prod_{m=1}^{|Q^{a}|}{\frac{(\tilde{N}^{ab}_{m})!(N^{ab}_{m}+|\Sigma^{b}|-1)!}{(\tilde{N}^{ab}_{m}+N^{ab}_{m}+|\Sigma^{b}|-1)!}}\\
                     \prod_{n=1}^{|\Sigma^{b}|}{\frac{(\tilde{N}^{ab}_{mn}+N^{ab}_{mn})!}{(\tilde{N}^{ab}_{mn})!(N^{ab}_{mn})!}}
\label{equProbOnline1}
\end{split}
\end{align}
where, $K$ is a proportional constant.

Thus, with Eq. \ref{equProbOnline1}, importance metrics of APs (i.e., when $a=b$) and RPs  (i.e., when $a\neq b$) are obtained with respect to the \textit{short subsequences}. In order to train the system-wide RBM, the metrics can be further normalized and converted to binary states ($0$ for low values and $1$ for high values) for APs and RPs. Note, from each subsequence, all the APs and RPs together form a binary vector of length $L = f^2$ ($L=\#AP+\#RP$, where $\#AP=f$, $\#RP=f \times (f-1)$). One such binary vector is treated as one training example for the system-wide RBM (with $f^2$ number of visible units) and many such examples are generated from different short subsequences extracted from the overall training sequence. Then a maximum likelihood method is used to train the RBM as mentioned in~\ref{sec:rbm}. Note, although in this paper we convert the $\Lambda^{i,j}$ metrics to binary values for the ease of RBM training, it is not mandatory for this training process.

\textbf{Remark}. With an optimized time lag in STPN learning, the STPN+RBM model can handle variable time lag situations and significantly reduces the complexity of the overall learning process, in comparison with other potentially spatiotemporal approaches, such as recurrent neural networks (RNNs).

\subsection{Anomaly detection process}\label{sec:anomaly}
The anomaly detection process developed here uses the concept of \textit{free energy} of RBM which is an energy based probabilistic graphical model. The energy function for an RBM is defined as
\begin{equation}
  E(\textbf{v}, \textbf{h})=-\textbf{h}^{T}\textbf{W}\textbf{v}-\textbf{b}^{T}\textbf{v}-\textbf{c}^{T}\textbf{h}
\end{equation}
where $\textbf{W}$ are the weights of the hidden units, $\textbf{b}$ and $\textbf{c}$ are the biases of the visible units and hidden units, respectively.

With the weights and biases of RBM, free energy can be obtained which is the energy that a single visible layer pattern would need to have in order to have the same probability as all of the configurations that contain $\textbf{v}$ \cite{hinton2012practical}
\begin{equation}
  e^{-F(\textbf{v})}=\sum_{\textbf{h}}{e^{-E(\textbf{v},\textbf{h})}}
\end{equation}
Another expression of free energy estimation is \cite{hinton2012practical}
\begin{equation}
F(v)=-\sum_{i}{v_{i}a_{i}}-\sum_{j}\log(1+e^{b_{j}+\sum_{i}{v_{i}w_{ij}}})
\label{equFreEngy}
\end{equation}

During training, weights and biases are obtained such that the training data has low energy. Therefore, during testing, an anomalous pattern should manifest itself as a low probability (high energy) configuration which can be used for anomaly detection. To detect an occurrence of anomaly, short testing subsequences are converted into an $f^2$-dimensional binary vector with the same \textit{inference phase} of the training process. Multiple testing (possibly overlapping) subsequences can be used to obtain a distribution of free energy. For the nominal condition, the distribution of free energy should be close to that of the training data, while the anomalous data should result in a different distribution of free energy (with increase in expected free energy).

\begin{figure*}[htb]
    \centering
    \subfigure[Nominal]{\includegraphics[scale=0.3]{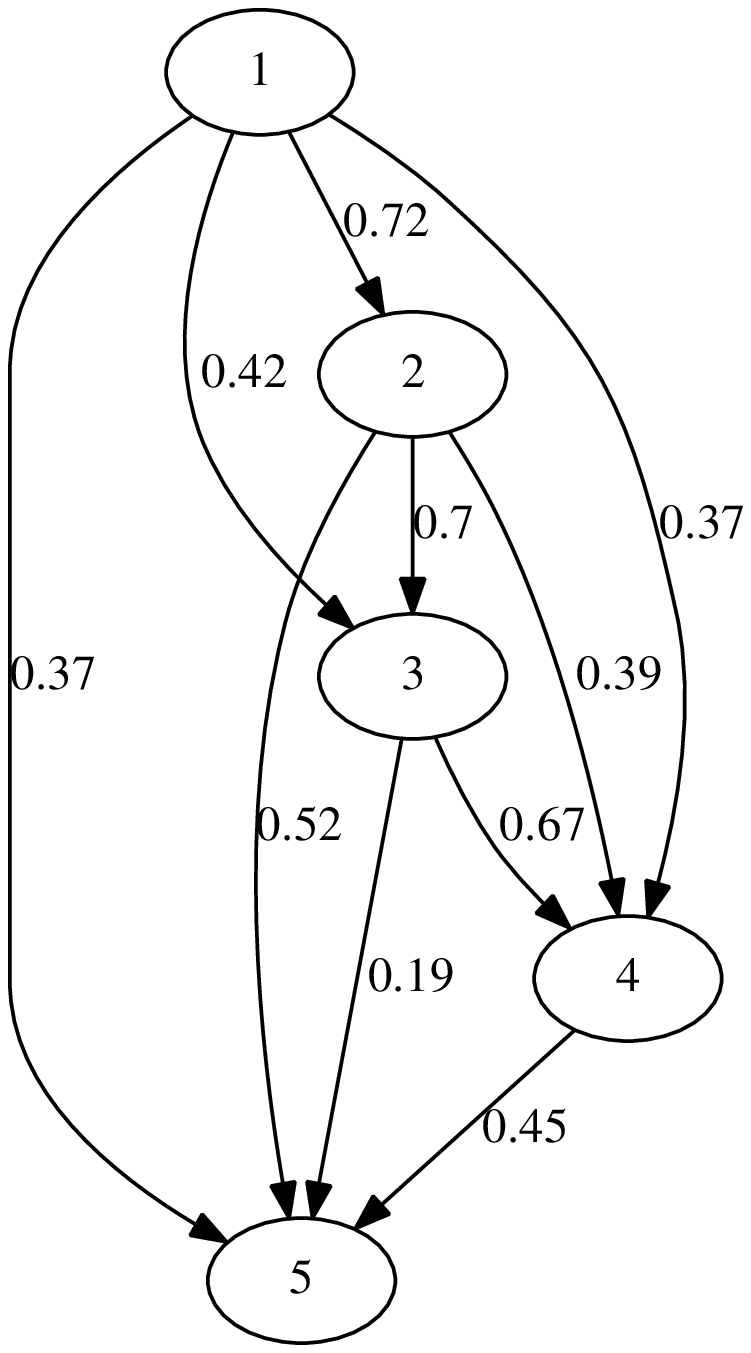}     }    
    \subfigure[Anomaly 1]{      \includegraphics[scale=0.35]{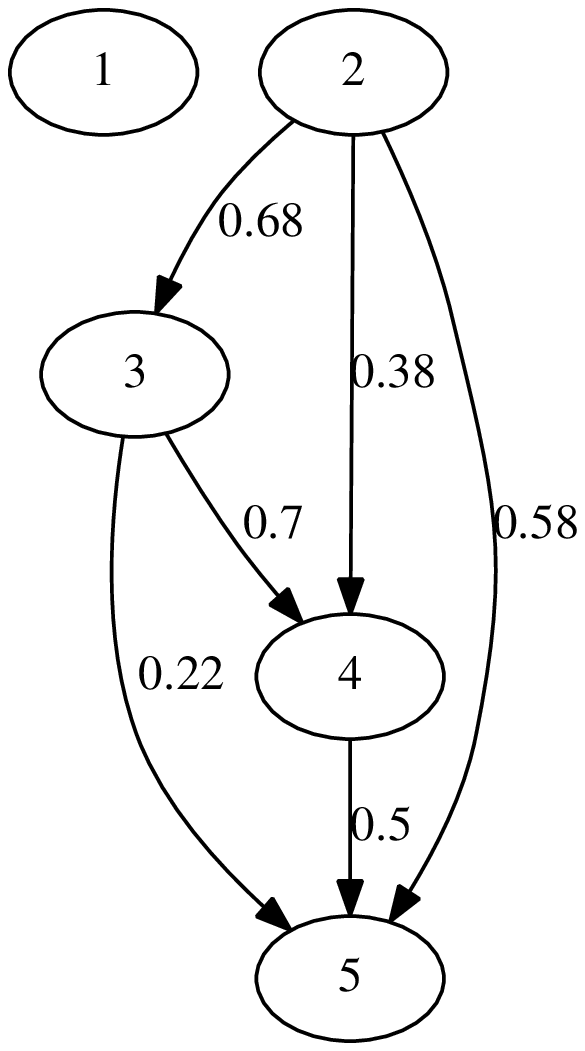}      }    
    \subfigure[Anomaly 2]{      \includegraphics[scale=0.35]{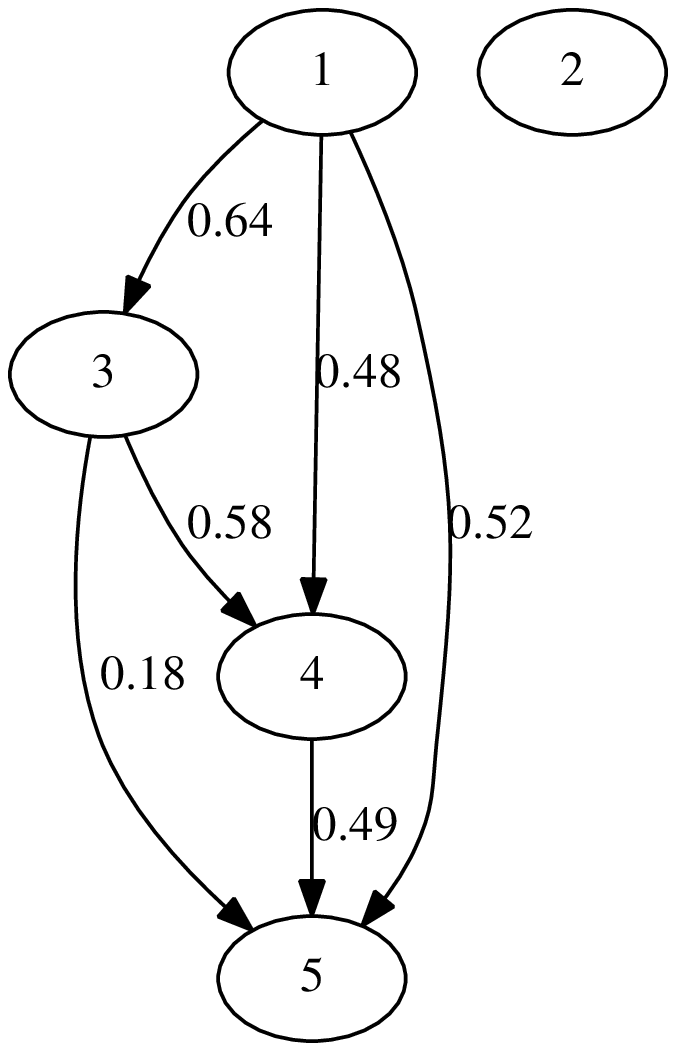}      }    
    \subfigure[Anomaly 3]{      \includegraphics[scale=0.35]{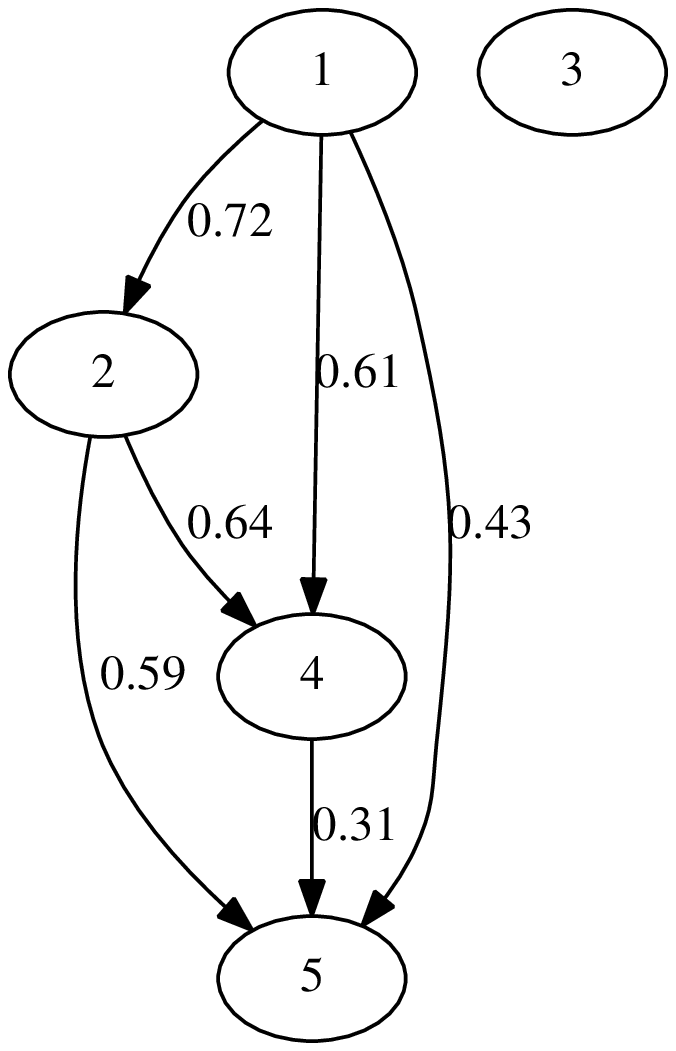}      }    
    \subfigure[Anomaly 4]{      \includegraphics[scale=0.35]{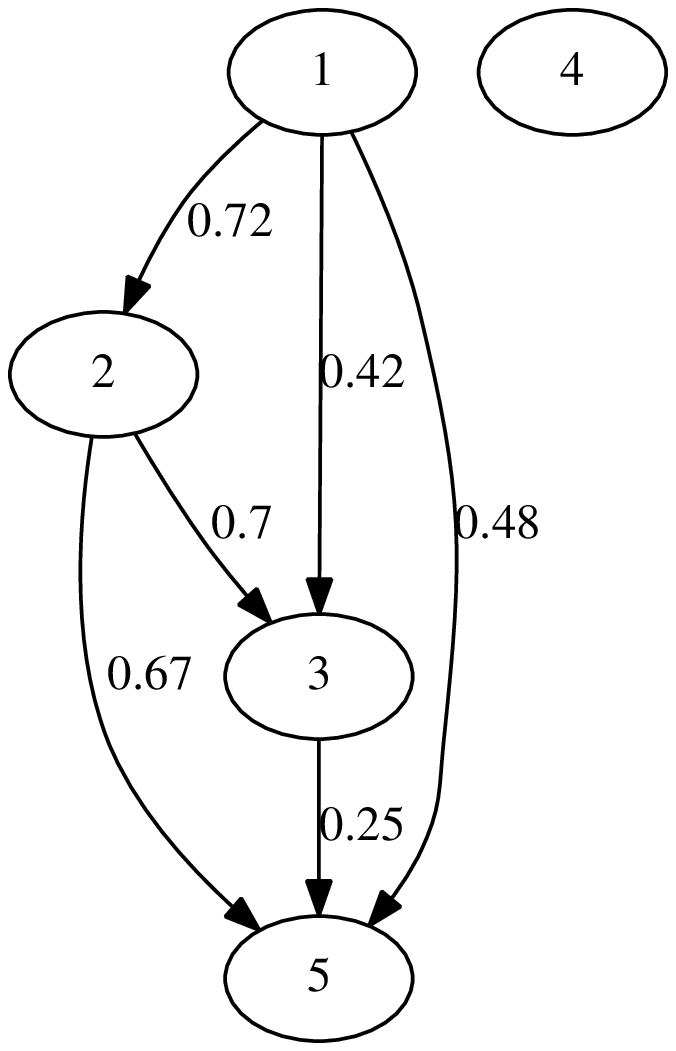}      }    
    \subfigure[Anomaly 5]{      \includegraphics[scale=0.35]{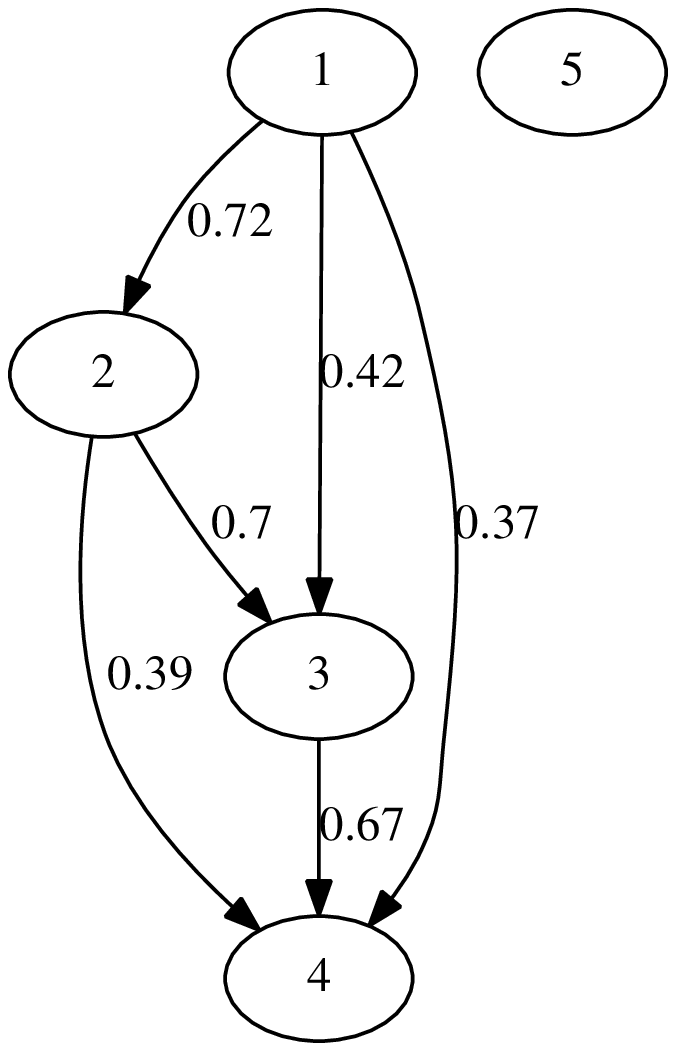}      } \vspace{-10pt}
    \caption{\emph{Graphical models defined for synthetic data generation and anomaly detection, where anomalous conditions occur due to failure of one vertex and the causal links to the failed vertex are lost}}
    \label{figSTPNanom5}\vspace{-10pt}
\end{figure*}

To compare the free energy distributions of testing data and training data, Kullback-Leibler divergence is considered \cite{kullback1951information}.

Since the Kullback-Leibler (KL) divergence is a non-symmetric information measure of distance between distribution $P$ and distribution $Q$, a symmetric Kullback-Leibler Distance (KLD)~\cite{kullback1951information} is used which is defined as,
\begin{equation}
  {KLD}(P||Q)=  {KL}(P||Q)+  {KL}(Q||P)
\label{equKLD}
\end{equation}

\section{Results and Discussions}
\label{secResults}
\subsection{Anomaly detection in simulation data}

\subsubsection{Synthetic data with Vector Autoregressive (VAR) Process}
\label{subsecCausual}
Vector autoregressive (VAR) process is widely applied in economics and other sciences as it is flexible and simple for multivariate time series data \cite{goebel2003investigating}. The basic model of VAR process $Y(t)={y_{i,t}, i=(1,2,\cdots,f),\ t \in \mathbb{N}}$ is defined as
\begin{equation}
y_{i,t}=\sum_{k=1}^{p}{\sum_{j=1}^{n}{A_{i,j}^{k}y_{j,t-k}+\mu_{t}}}, \quad j=(1,2,\cdots,f)
\end{equation}
where $p$ is time lag order, $A_{i,j}$ is the coefficient of the influence on $i$th time series caused by $j$th time series, and $\mu_{t}$ is an error process with the assumption of white noise with zero mean, that is $E(\mu_{t})=0$, the covariance matrix, $E(\mu_{t}\mu_{t}^{\prime})=\Sigma_{\mu}$ is time invariant.

Synthetic multivariate data is generated using the VAR process in this study. A hierarchical structure of five-vertex network is considered with different interactions among vertices. Various interaction patterns represent nominal and off-nominal conditions in a distributed CPS. Data for two case studies are generated: \textit{Case I:} six patterns are defined, where the first one is considered as nominal condition and the others as anomalous; \textit{Case II:} eight patterns are defined, where the first three are considered as nominal conditions, and remaining five as anomalous. The second case is to simulate the CPS scenario with multiple nominal operating modes.

\begin{figure}[h]
  \centering
  \includegraphics[scale=0.55]{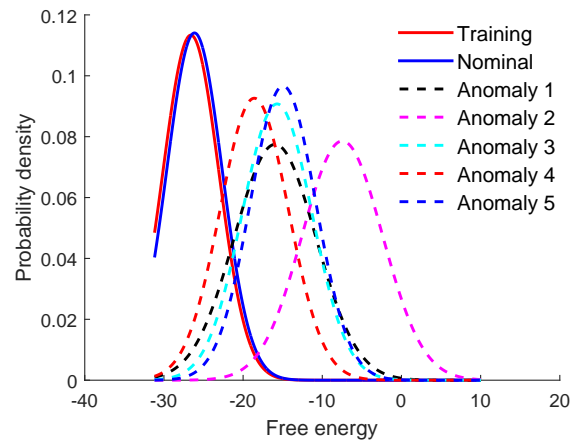}   \vspace{-8pt}
  \caption{\emph{Distribution of free energy of RBM with multiple testing samples from each nominal/anomalous condition}}
  \label{figAnomDet5}\vspace{-10pt}
\end{figure}

\begin{figure*}[t]
    \centering
    \subfigure[Nomina 1, 2, and 3]{\includegraphics[scale=0.3]{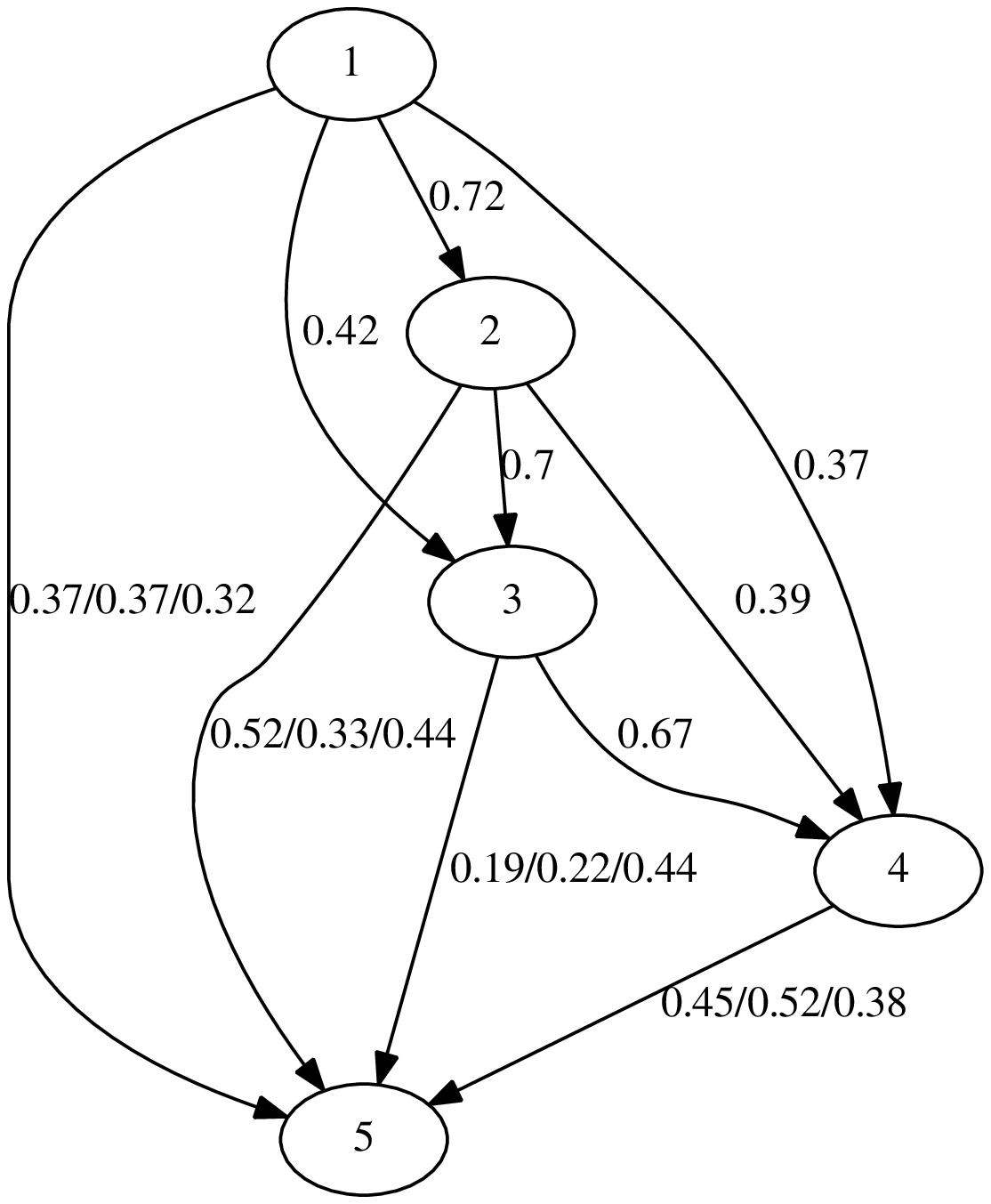}    }    
    \subfigure[Anomaly 1]{          \includegraphics[scale=0.32]{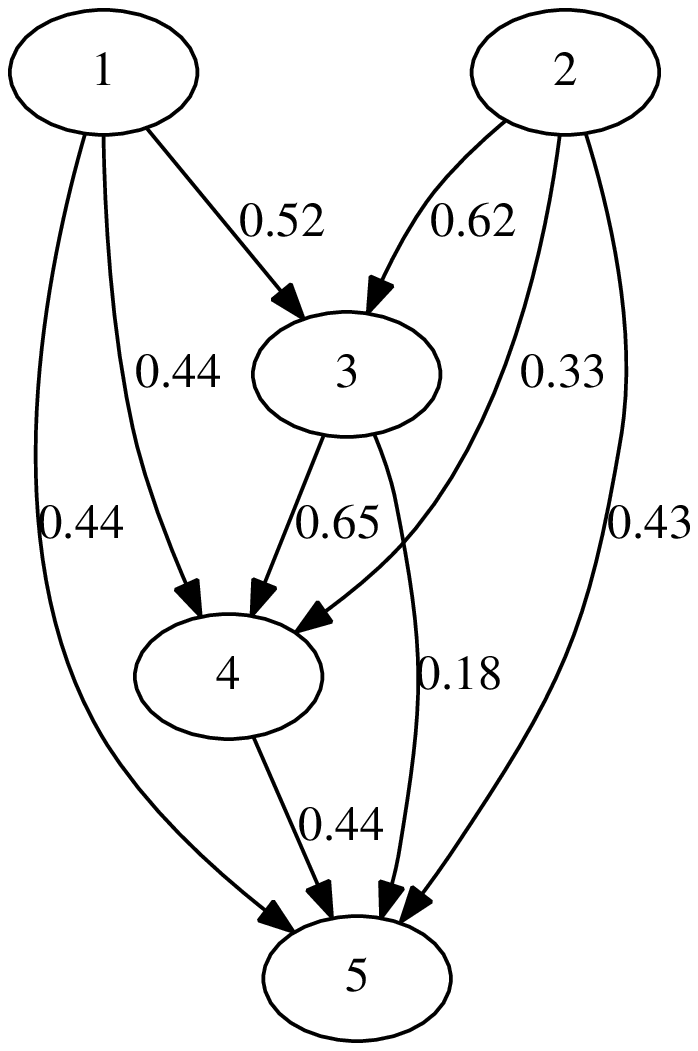}     }    
    \subfigure[Anomaly 2]{          \includegraphics[scale=0.3]{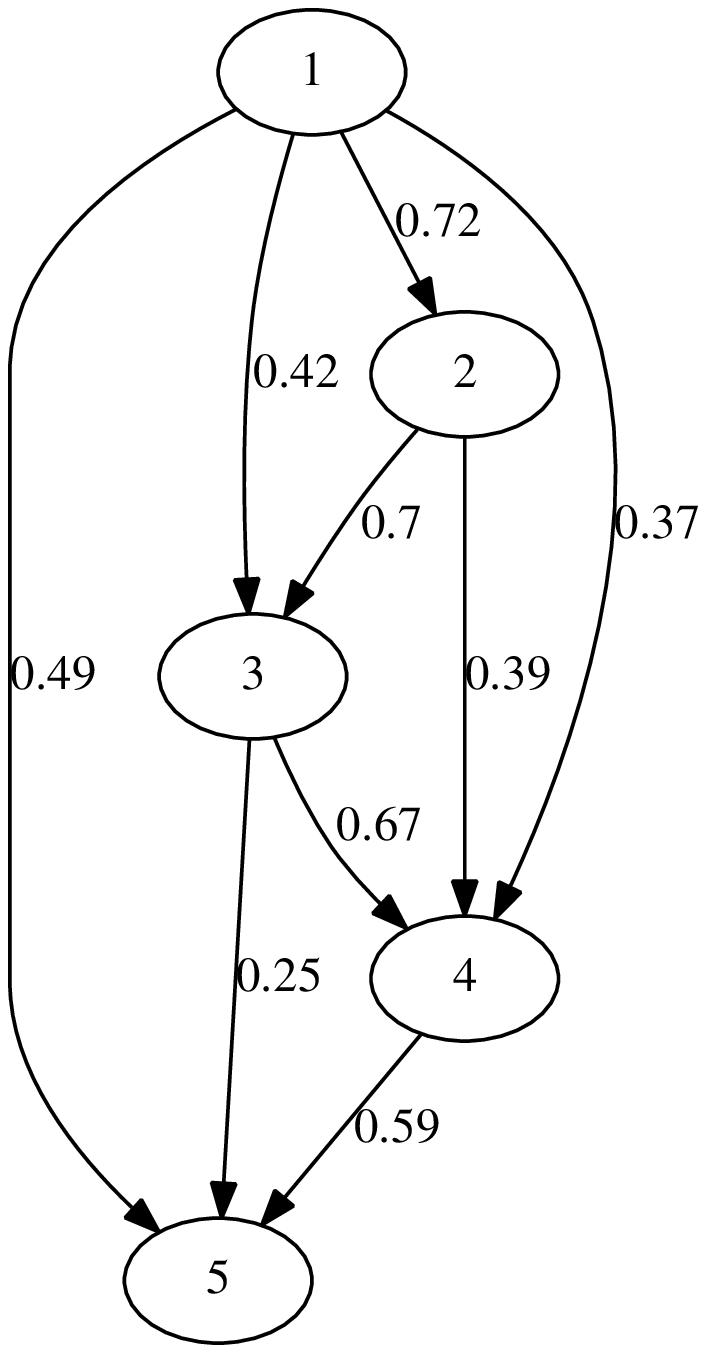}      }   
    \subfigure[Anomaly 3]{          \includegraphics[scale=0.32]{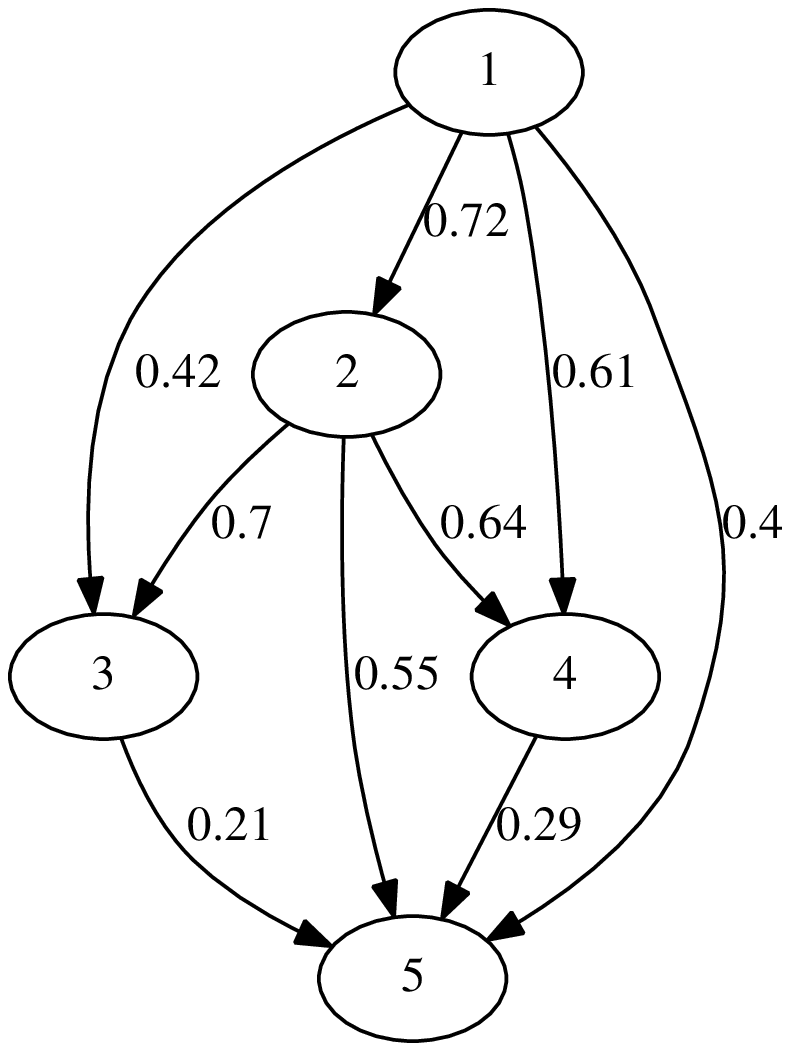}      }   
    \subfigure[Anomaly 4]{          \includegraphics[scale=0.32]{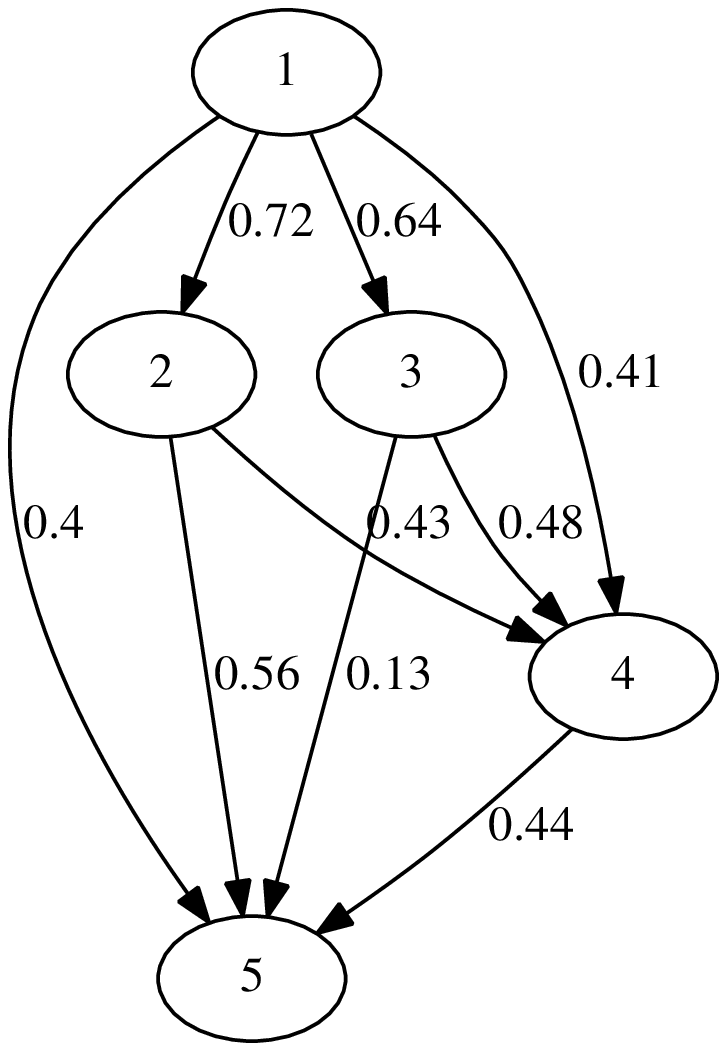}      }   
    \subfigure[Anomaly 5]{          \includegraphics[scale=0.32]{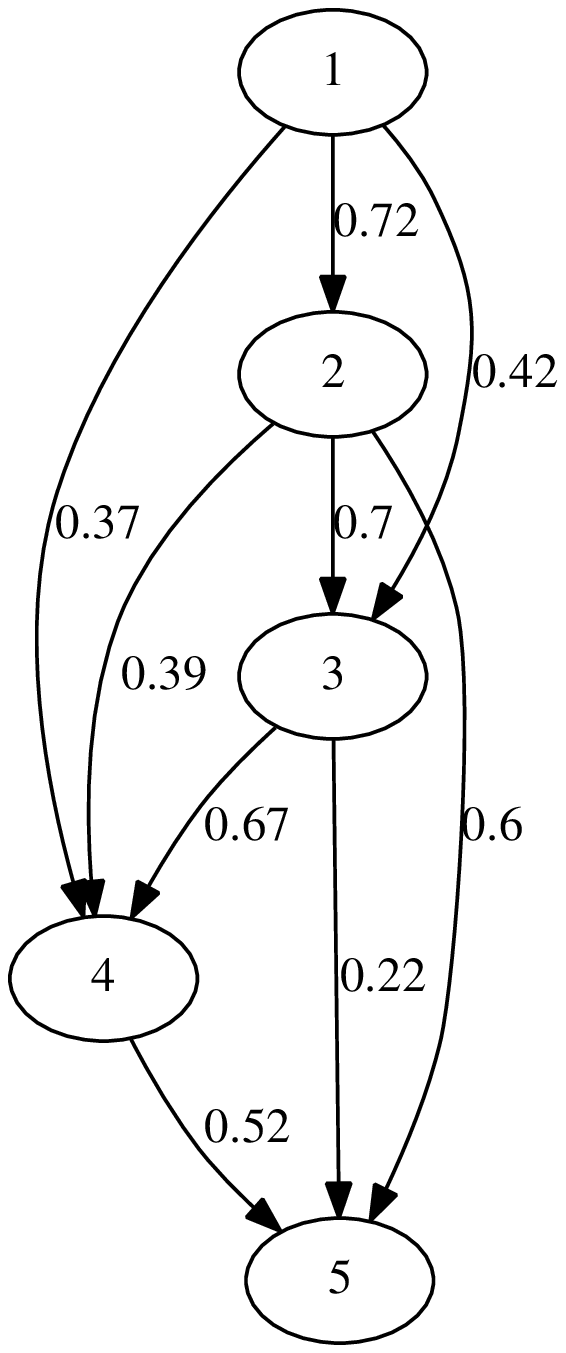}      } \vspace{-10pt}
    \caption{\emph{Graphical models defined for synthetic data generation and anomaly detection, where the nominal condition consists of (first) three graphs representing three different operation modes, and the others are considered as anomalous}}
    \label{figSTPNanom8}\vspace{-10pt}
\end{figure*}

STPNs are learnt from raw data generated by the VAR models representing different interaction patterns. For unsupervised RBM learning, only nominal patterns are used to obtain the weights and biases. Data from all patterns are used with the trained RBM to compute the free energy with respect to the different pattern inputs. Evaluation of free energy is performed with Gaussian assumption of multiple testing inputs.

\subsubsection{Case Study I: Single nominal mode}
\label{subsubsecSngAnomDet}
While the details of STPN learning is skipped for brevity, the predefined graphical models are shown in Fig.~\ref{figSTPNanom5}. Atomic and relational patterns of the nominal condition are used for training the RBM. Testing data corresponding to all 6 patterns are used to generate the probability distribution of free energy with respect to the trained RBM and the results are shown in Fig.~\ref{figAnomDet5}. It is evident, while the free energy distribution of the nominal pattern matches quite closely with that of the training patterns, free energy distributions of other patterns are quite different and moves to the right. This means such patterns have high free energy and hence low probability of occurrence. Quantitatively, the KLD metrics between the free energy distribution of the training/nominal pattern and those of all the other test patterns are 0.01, 6.82, 21.70, 7.63, 4.17, and 9.08 respectively.

\begin{figure}[htbp]
  \centering
  \includegraphics[scale=0.55]{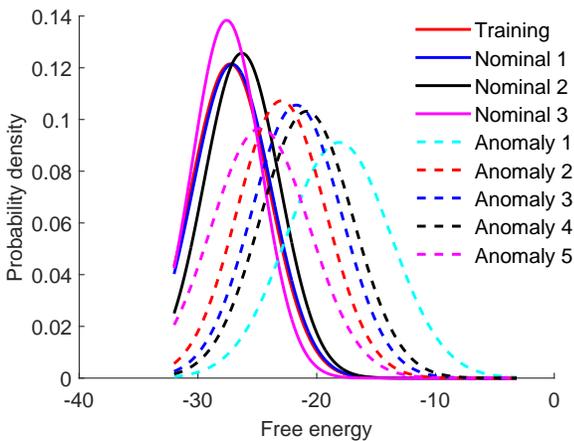}  \vspace{-8pt}
  \caption{\emph{Distribution of free energy of RBM with multiple testing samples for each nominal/anomalous condition}}
  \label{figAnomDet8}\vspace{-10pt}
\end{figure}

\subsubsection{Case Study II: Multiple nominal modes}
\label{subsubsecMultiAnomDet}

Similar to the previous case, the predefined graphical models are shown in Fig.~\ref{figSTPNanom8} and the free energy distributions are shown in Fig.~\ref{figAnomDet8}. As expected, free energy distributions for first three patterns (Nominal 1, 2 and 3) are similar to that of the training data with KLD values 0.002, 0.068 and 0.040 respectively. The figure also shows that anomalous patterns can be clearly identified for conditions 4 through 8 (with KLD values 6.067, 1.461, 2.420, 3.199, and 0.906 respectively). Overall, the result clearly shows that the proposed framework can capture multiple nominal behaviors within one model while slight change in causality patterns can be detected efficiently.

\begin{figure*}[htbp]
  \centering
  \includegraphics[scale=0.5]{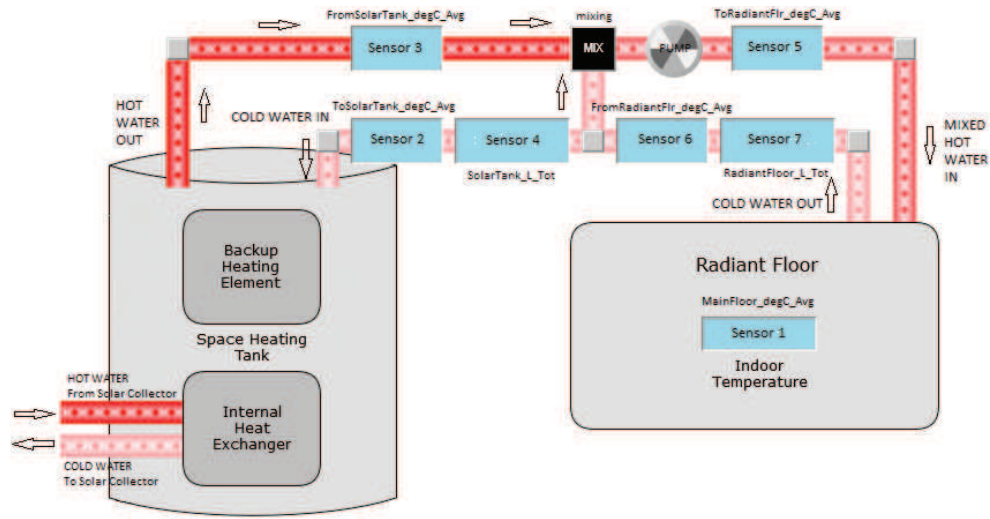}
  \caption{\emph{Radiant floor heating subsysem within the integrated building system, including 7 sensors measuring temperature, flow rate, and power consumption. Sensors 1, 2, 3, 5, and 6 are temperature sensors, sensors 4 and 7 are flow rate sensors, sensors 8 and 9 measure power consumption of the heating tank (not marked in the figure).}}
  \label{figInterlock}\vspace{-10pt}
\end{figure*}
\subsection{Validation on a Real System - Integrated Building}
\label{subsecRead}
The Interlock House, an Iowa NSF EPSCOR (experimental program to stimulate competitive research) (\# EPS-1101284.) community lab, located at the Honey Creek Resort State Park, Iowa is a smart home designed to ``interlock'' with its environment, its occupants, and its active and passive energy systems. The goal of this test bed is to perform as a net-zero energy house that (ideally) produces as much energy as it consumes. With all its intelligent sensing and control systems, it can be classified as a typical cyber-physical system, with collaborating computational elements controlling the interconnected physical sub-systems. One of the key systems for building operation is the HVAC (Heating, Ventilation \& Air-Conditioning) system that is comprised of the following three independent sub-systems: (a) Radiant Floor Heating (Space Heating), (b) Domestic Water Heating, and (c) Space Cooling and Ventilation. For this paper, the Radiant Floor Heating subsystem is chosen and studied. A schematic of the Radiant Floor Heating system is shown in Fig. \ref{figInterlock}.

\begin{figure*}[htbp]
    \centering
    \subfigure[sensor 3 in winter]{\includegraphics[scale=0.55]{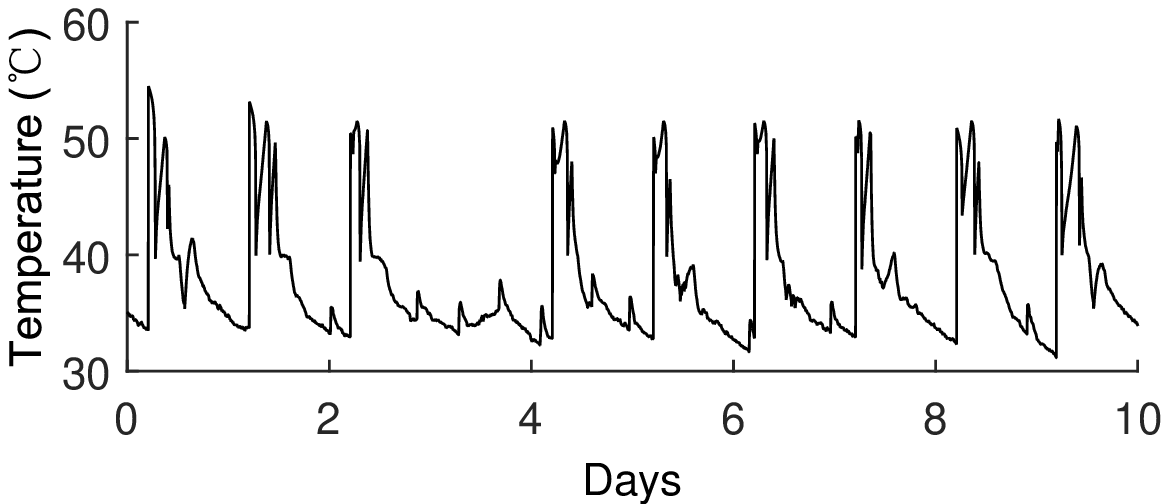}    }
    \subfigure[sensor 7 in winter]{\includegraphics[scale=0.55]{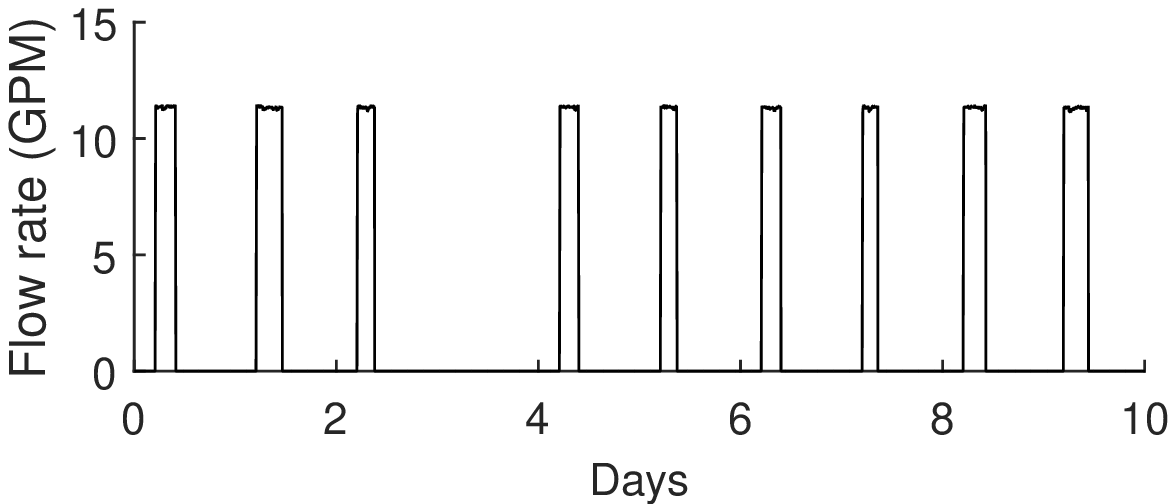}     } \\ \vspace{-10pt}
    \subfigure[sensor 3 in summer]{\includegraphics[scale=0.55]{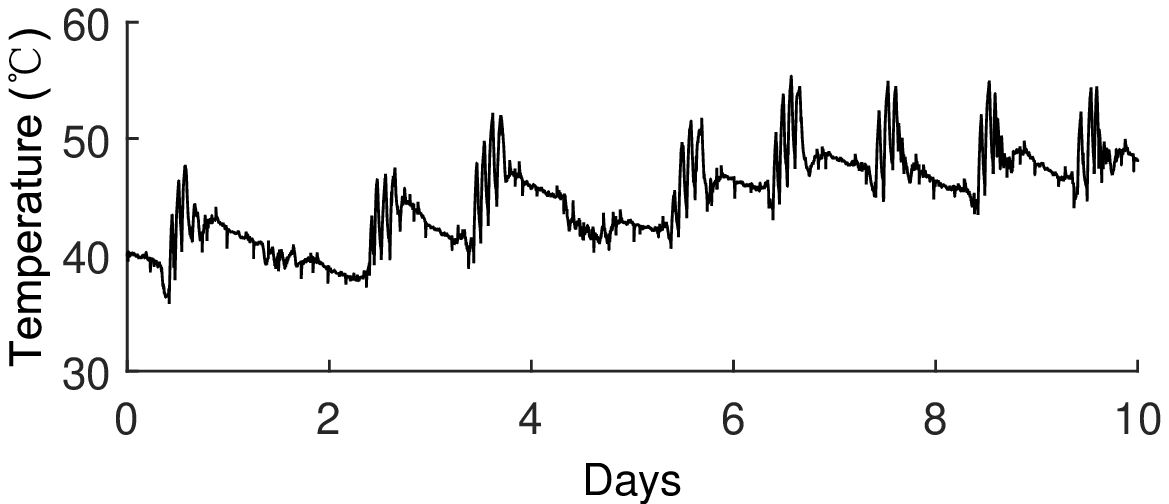}    }
    \subfigure[sensor 7 in summer]{\includegraphics[scale=0.55]{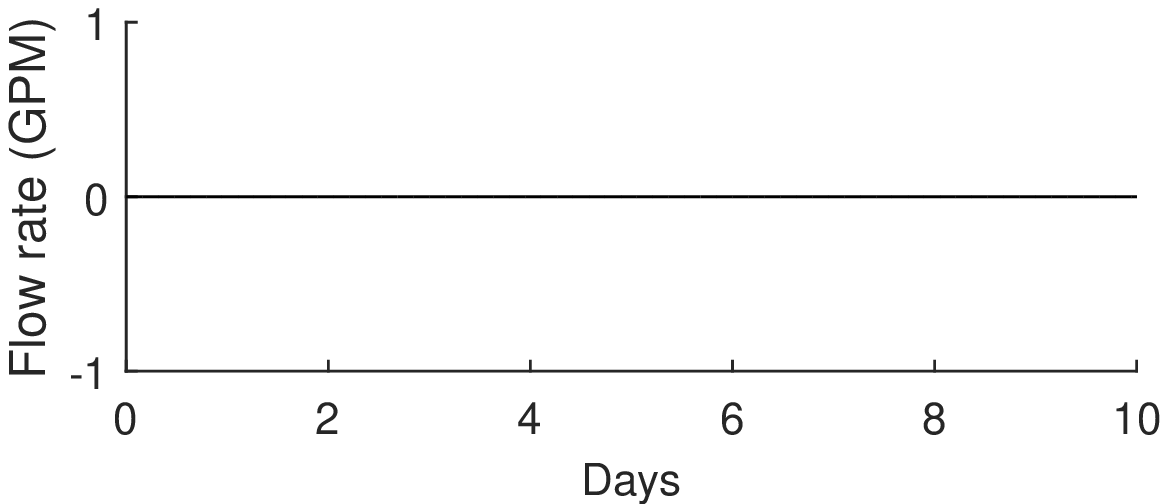}     } \vspace{-10pt}
    \caption{\emph{Data collected at the Interlock House in winter and summer, the subsystem includes 9 sensors measuring various temperatures, flow rates, and power consumptions.
    }}
    \label{figTestbedDataNor}\vspace{-10pt}
\end{figure*}
\begin{figure}[h]
    \centering
    \subfigure[sensor 3]{\includegraphics[scale=0.35]{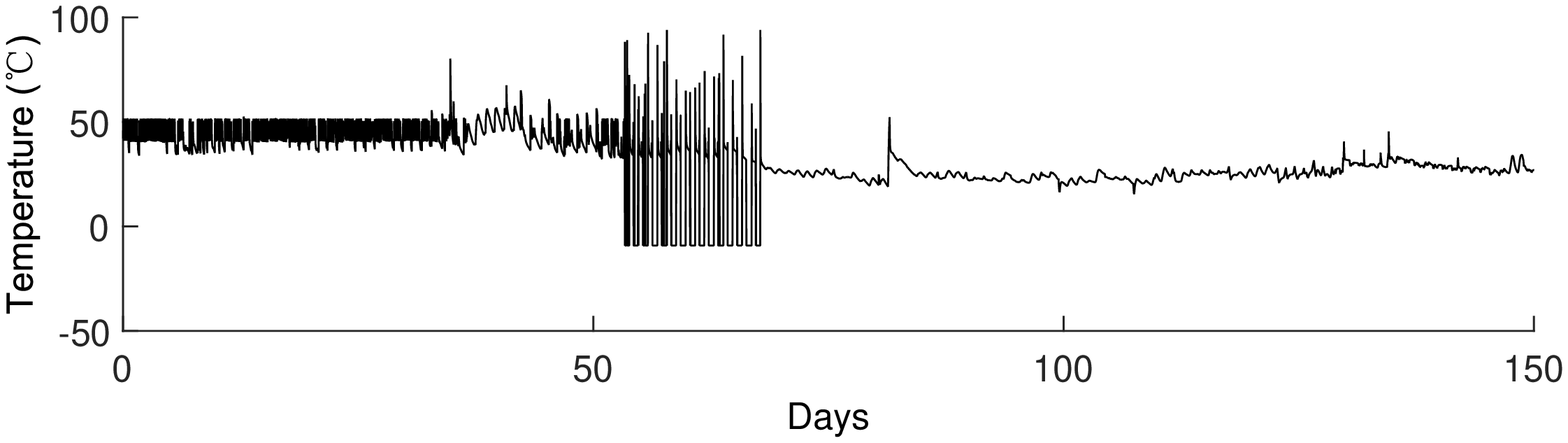}    } \\ \vspace{-10pt}
    \subfigure[sensor 7]{\includegraphics[scale=0.35]{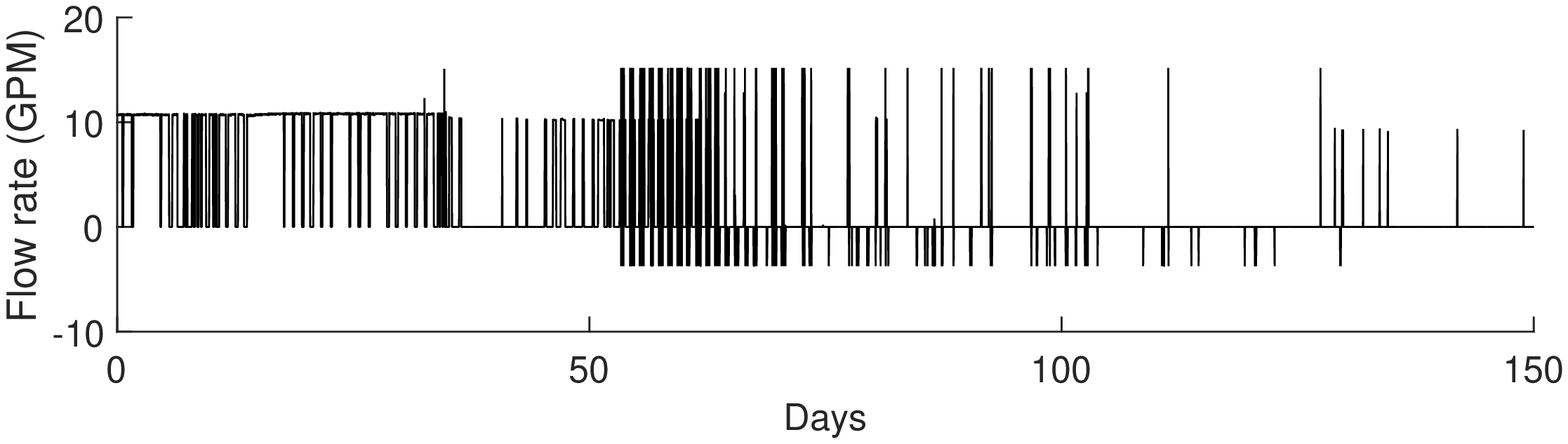}     } \vspace{-10pt}
    \caption{\emph{Data collected at the Interlock House during Feb., 2015 -- Jun., 2015.
    }}
    \label{figTestbedDataAnom}\vspace{-10pt}
\end{figure}

Sensor data related to its operation is collected for a period of 1 year from Jul., 2014 to Jun., 2015. Within this time-frame, data collected for two different seasonal operational modes (viz. winter and summer modes) is studied. For winter mode of operation, the month of Nov., 2014 is chosen and Jul., 2015 is chosen for summer mode. Typical data for two representative sensors are shown in Fig.~\ref{figTestbedDataNor}, where features vary in seasons and are used for unsupervised learning. Another timeframe from Feb., 2015 to Jun., 2015 is considered for validation of the proposed approach that covers a wide range of behavior over both winter and summer, as shown in Fig.~\ref{figTestbedDataAnom}.

\textit{Winter Mode study (Nov., 2014):} In Winter, periodic heating of the radiant floor space is required. The heating is set to start every morning at 5:00 am if the indoor temperature at that point is below 19.5 deg C. The heating turns off when the heating setpoint (plus the dead band) is reached, which was around 20 deg C, with the dead band being 1 deg C (+0.5C/-0.5C).
\begin{itemize}
  \item This is evident from Fig.8 (a) where, we see a rise in the temperature of the water supplied from the space heating tank to radiant floor at approximately every 24 hour time interval, when demand for heating is high. The temperature of the supplied water gradually decreases as demand for heating decreases and it rises again at 5:00 am next day, if the set conditions are met. \vspace{-5pt}
  \item Consequently, it is seen in Fig.8 (b) that the flowmeter measuring flow of water returning from the radiant floor records flow of water in a similar pattern. Flow measurements are recorded at every 24 hour interval when heating is required, depending on the set conditions.  \vspace{-5pt}
  \item It is observed that on the fourth day, flow measurements remained zero and floor heating was not carried out. This is because, the indoor temperature that day remained above 19.5 deg C at 5 am.  \vspace{-5pt}
\end{itemize}
	
\textit{Summer Mode study (July, 2014):} In Summer, water is heated in the Space Heating Tank periodically, as is evident from Fig. 8 (c), but as floor heating in summer is not required, supply to the radiant floor remains 0 all throughout, as is evident from Fig. 8 (d).

\textit{Anomaly Detection:} On the study conducted on the data from Feb., 2015 -- Jun., 2015, we notice that anomalies creep in certain sensor observations (see Fig.~\ref{figTestbedDataAnom}). While the first anomaly starts around March 26, 2015 (day 55 in the plot) and continues till June with erroneous readings in several sensors, a second anomaly appears during June, 2015. However, careful observation shows that the second anomaly is extremely intermittent and only manifests itself in a few number of (three out of nine) sensor measurements.

\begin{figure}[h]
  \centering
  \includegraphics[scale=0.28]{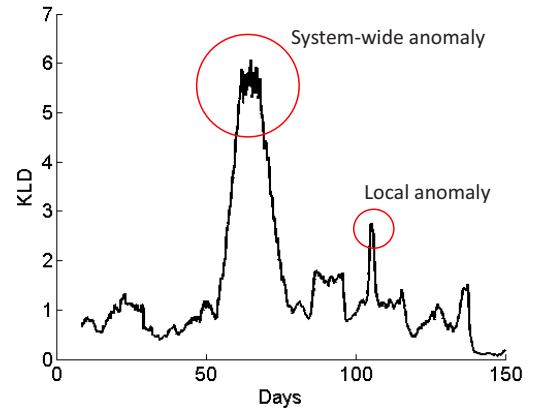}
  \caption{\emph{Online anomaly detection during Feb. 2015 -- Jun. 2015.  }}
  \label{figTestbedDataAnomDet}\vspace{-15pt}
\end{figure}

Online anomaly detection is carried out using the proposed technique and the results are shown in Fig.~\ref{figTestbedDataAnomDet}. The algorithm detects the first anomaly quite easily that appears as a large peak in the plot. The second anomaly which is local, intermittent and not much pronounced, is also detected and appears as the second sharp peak in the plot. Another crucial observation is that we use different sets of test data for two different modes of operation (summer and winter). The proposed technique can effectively handle nominal data for all such modes without false detection.

\subsection{Discussions}
Anomalies in distributed CPSs vary in mechanisms, features, and duration, and this makes anomaly detection difficult, especially the collection of labeled data for all possible anomalies. The proposed framework in this work only needs nominal data, and anomaly detection is considered as a low probability event conditioned on the nominal data. The results show that the free energy distribution during an anomalous condition is different from that of the nominal condition, and a metric such as KLD can be used to quantify the change. Furthermore, it has the potential for monitoring all the way from small physical degradation to severe faults or cyber attacks, where a small KLD signifies slight change in the causal pattern (STPN) that possibly indicates early-stage degradation, localized cyber anomaly or precursor of a fault.

\textbf{Multiple nominal modes:} The proposed framework can capture multiple modes as nominal condition (e.g., Fig. \ref{figSTPNanom8} and Fig. \ref{figTestbedDataNor}), within a single model and hence reduces the modeling and reasoning complexity. Traditionally, a graphical model such as STPN can represent one operating mode at a time, and multiple STPNs are required to represent various operating modes. Moreover, it is difficult to determine the number of STPNs needed for representing all nominal modes, as a small number of changes in the patterns can result in a larger number of combinations, which may cause dimension explosion and extreme difficulty in real life applications.

\textbf{Local and Global anomalies:} In the simulated cases shown in Fig.~\ref{figSTPNanom8}, only one causal connection is removed (e.g., arrow $1 \to 2$ is disconnected in STPN 4, comparing with STPNs 1, 2, and 3), and it represents a local anomaly. In the real case study, most of the measurements are abnormal during days 55-70 in Fig. \ref{figTestbedDataAnom}, and it is a type of global anomaly. For both of types of anomalies, the proposed approach can distinguish them from the nominal modes.

\textbf{Mixed data types:} In the real case study (Section \ref{subsecRead}), mixed data types are used. For example, temperature measurement in Fig. \ref{figTestbedDataNor} (a) is a representative continuous sensor observation. On the other hand, the flow rate measurement in Fig.~\ref{figTestbedDataNor} (b) appears as fairly discrete due to the on-off control schedule. The proposed approach is able to extract features from both types of data and fuse them into one model. This is a critical need for a distributed CPS with continuous temporal information from sensors and actuators of the physical space and discrete event-driven data from the cyber space.

\textbf{Robustness:} Although detailed false alarm and missed detection studies will be done in near future, from a qualitative standpoint, the approach is seen to be quite robust as it is designed to identify only persistent anomalies. This property is inherent as the anomaly detection is performed with short subsequences and not based on individual data points. As a consequence, no false alarms are observed when only a few sensors show bad readings for a very short time window. For example, error in readings occurs occasionally during the first 50 days (in Fig.~\ref{figTestbedDataAnom}), where no anomaly is detected during that period (in Fig. \ref{figTestbedDataAnomDet}).

\textbf{Remark:} The proposed framework is categorized as unsupervised because unlike typical diagnostic frameworks, no labeled training data for various fault types and fault locations are required; only assumption made for the training data is that it represents nominal condition(s) with a high probability. Even with such limited training data, the proposed tool has the potential of identifying root-cause of a detected anomaly. Demonstration of such a capability remains a critical future work. Furthermore, for learning multiple STPNs with one-layer RBM, the results are acceptable in the above cases. However, differences among nominal modes may be quite large in some cases and one-layer RBM may not capture all the features in such cases. To address this issue, stacked RBM (i.e., a deep belief network (DBN)) can be used \cite{RB08,SH12} to learn more complex features.

\section{Conclusions}
\label{secconclusions}
This work presents a new data-driven framework for system-wide anomaly detection to address a wide-range of nominal operating modes and unforeseen anomalous situations without comprehensive labeled training data in distributed CPSs. The framework involves a spatiotemporal feature extraction scheme for discovering and representing causal interactions among the subsystems of a CPS, and a free energy estimation of system-wide patterns using an RBM. The results show that the proposed framework can capture multiple diverse nominal modes within a single probabilistic graphical model, and detect anomalies via identifying a low probability event. Multiple case studies with simulation and real data validate accuracy, robustness, ability to handle mixed data type and adaptiveness (i.e., local vs. global) of the proposed method.

While the current work is focusing on validating the method for a large variety of scenarios, quantifying false alarm and missed detection rates, further works will pursue the following: (i) using the graphical model for root-cause analysis for various anomalies, (ii) stacked RBM approach to capture more complex nominal patterns and (iii) detection of simultaneous multiple faults in distributed CPS.

\vspace{-5pt}
\section*{Acknowledgments}
The authors would like to express their sincere gratitude to Dr. Soumalya Sarkar and Mr. Adedotun Akintayo for their thoughtful comments on the framework, Prof. Ulrike Passe, Ms. Shan He (Center for Building Energy Research at Iowa State University), and Mr. Mike Wassmer (Live to Zero Inc.) for their support in the validation process using real data, and for providing access to their historic data collected as part of the building science plank research of the Iowa NSF EPSCOR project funded under grant number EPS-1101284. \vspace{-5pt}

\bibliographystyle{abbrv}
\bibliography{ICCPS2016}
\end{document}